\documentclass[letterpaper]{article} 
\usepackage{aaai2026}  
\usepackage{times}  
\usepackage{helvet}  
\usepackage{courier}  
\usepackage[hyphens]{url}  
\usepackage{graphicx} 
\usepackage{natbib}  
\usepackage{caption} 
\usepackage{algorithm}
\usepackage{algorithmic}
\usepackage{tabularx}
\usepackage{array}
\usepackage{amsmath}
\usepackage{graphicx}
\usepackage{url}
\usepackage{booktabs}
\usepackage{xcolor}
\usepackage{multirow}
\usepackage{makecell}
\usepackage{threeparttable}
\usepackage{amssymb}
\usepackage{bm}
\usepackage{seqsplit}
\usepackage{float}
\usepackage{subfigure}
\usepackage{amsmath}
\usepackage{inconsolata}
\newcommand{\vx}{\bm{x}}
\newcommand{\ve}{\bm{e}}

\usepackage{newfloat}
\usepackage{listings}
\DeclareCaptionStyle{ruled}{labelfont=normalfont,labelsep=colon,strut=off} 
\floatstyle{ruled}
\newfloat{listing}{tb}{lst}{}
\floatname{listing}{Listing}
\nocopyright

\title{Discovering Mathematical Equations with Diffusion Language Model}

\author{
     Xiaoxu Han\textsuperscript{\rm 1},  Chengzhen Ning\textsuperscript{\rm 1}, Jinghui Zhong\textsuperscript{\rm 1}, \\
     Fubiao Yang\textsuperscript{\rm 1}, Yu Wang\textsuperscript{\rm 2}, Xin Mu\textsuperscript{\rm 2}
}
\affiliations{
    \textsuperscript{\rm 1} South China University of Technology \\

    \textsuperscript{\rm 2} Pengcheng Laboratory\\

}

\usepackage{bibentry}

\begin{document}

\maketitle

\begin{abstract}
Discovering valid and meaningful mathematical equations from observed data plays a crucial role in scientific discovery.
While this task, symbolic regression, remains challenging due to the vast search space and the trade-off between accuracy and complexity.
In this paper, we introduce DiffuSR, a pre-training framework for symbolic regression built upon a continuous-state diffusion language model. 
DiffuSR employs a trainable embedding layer within the diffusion process to map discrete mathematical symbols into a continuous latent space, modeling equation distributions effectively. 
Through iterative denoising, DiffuSR converts an initial noisy sequence into a symbolic equation, guided by numerical data injected via a cross-attention mechanism.
We also design an effective inference strategy to enhance the accuracy of the diffusion-based equation generator, which injects logit priors into genetic programming.
Experimental results on standard symbolic regression benchmarks demonstrate that  DiffuSR achieves competitive performance with state-of-the-art autoregressive methods and generates more interpretable and diverse mathematical expressions.

\end{abstract}

\section{Introduction}

Mathematical equation synthesis, essential for computations and proof construction~\cite{chudnovsky2020computer}, is challenging due to nested structures, context-sensitive symbols, and strict grammars.
Goal-directed equation synthesis tasks further increase complexity.
A well-known example is symbolic regression~(SR)~\cite{la2021contemporary,kahlmeyer2025dimension}, which plays a crucial role in scientific discovery across domains such as clinical prediction~\cite{la2023flexible} and physical law identification~\cite{udrescu2020ai}.
SR aims to find equations from observed numeric data, and has been proven to be an NP-hard problem~\cite{virgolin2022symbolic}, due to its vast search spaces and trade-offs among accuracy, complexity, and diversity. 
The traditional dominant technique for SR is Genetic Programming~(GP)~\cite{koza1994genetic}, which benefits from a robust global search capability. Whereas, searching randomly from scratch, GP suffers from low search efficiency and overly high program complexity~\cite{han2025transformer}.

Deep generative models have been applied to equation synthesis and SR, achieving significant performance~\cite{reddy2025towards}.
A common generative approach is to leverage the variational autoencoders~(VAEs) to model the distribution of mathematical equations~\cite{dai2018syntax,mevznar2023efficient}. 
Early research encoded equations as token sequences with recurrent models and decoded them from a continuous latent space~\cite{gomez2018automatic}.
 Subsequent work imposed stronger syntactic priors, either by encoding context-free grammar production rules \cite{dai2018syntax,kusner2017grammar} or by traversing expression trees \cite{mevznar2023efficient}. 
VAE-based methods for equation synthesis can be extended to solve SR, yet they struggle with effectively incorporating conditional guidance (e.g., numerical data) during training. 
To adapt pre-trained VAE models for SR, gradient-free search methods, such as evolutionary search, are employed~\cite{kusner2017grammar,gomez2018automatic,mevznar2023efficient}. 
A key issue is that undirected search leads to low search efficiency, with the necessity for numerous inferences.
Besides, random sampling from the latent space can generate invalid or nonsensical expressions, as some regions may not correspond to grammar correct expressions.
Recently, autoregressive models have emerged as an alternative approach for solving SR in an end-to-end manner~\cite{biggio2021neural,kamienny2022end}. Based on large-scale pre-training, they exhibit robust generalization performance on unseen SR tasks. However, the autoregressive nature introduces intrinsic limitations~\cite{li2022diffusion,nie2025large}.
The fixed generation order restricts their adaptability, especially in contexts requiring precise control over the generation process.
Additionally, the token-by-token generation prevents parallel execution during inference, limiting efficiency improvements.

Treating generation as an iterative denoising process, diffusion models~\cite{ho2020denoising} eliminate the aforementioned constraints in autoregressive models and inject conditions at every denoising step.
Inspired by the success of diffusion models in domains like image and text generation~\cite{saharia2022photorealistic,li2022diffusion,gong2024text,wang2025stay,nie2025large}, we aim to explore the potential of using diffusion models to overcome the challenges of equation synthesis and SR.
In this paper, we introduce a novel diffusion-based framework for SR, \textbf{DiffuSR}. 
To model the distribution of mathematical equations, we adapt the diffusion process to operate in the discrete token space.
Specifically, a diffusion language model is leveraged to collectively and iteratively update token embeddings within equation sequences.
DiffuSR can be viewed as a numerical-to-symbolic framework, which consists of three key components: a numerical data processing embedder, a numerical data points encoder, and a discrete diffusion-based equation generator. The diffusion-based generator would iteratively denoise a sequence of Gaussian vectors into word vectors, and the control information is injected by running cross-attention on intermediate latent variables.
All symbolic tokens are predicted simultaneously, emitting a logit matrix that provides the probability of every token at every position. 
The emitted logit matrix is leveraged to guide the GP search by seeding the greedy decoded solution and guiding the subtree built up through sampling from the row, which provides a way to correct and improve the greedy solution. 
Experimental evaluations on SR benchmark datasets demonstrate that our method surpasses the state-of-the-art~(SOTA) autoregressive models in both accuracy and interpretability.
Moreover, we also show that DiffuSR tends to generate more diverse and human-readable equations, an essential capability for practical applications.
Additionally, we highlight the importance of diversity in generation, which enhances the model’s ability to recover and understand the underlying black-box formula.

Our main contributions are as follows:
\begin{itemize}
\item[$\bullet$]  We propose DiffuSR, introducing a continuous-state diffusion language model to the numerical data-guided equation generation task, SR, offering 
a novel approach
for concise and precise equation generation.



\item[$\bullet$] To further refine the obtained greedy solution, decoding strategies are designed. 
We propose a diffusion-guided GP search method, where the diffusion logits act as a flexible, position-wise structural prior that GP can incorporate at any depth in its architecture.

\item[$\bullet$] 
We conduct comprehensive evaluations using a variety of SR benchmark datasets.
The empirical results show that DiffuSR can find diverse, human-readable, and high-fitting accuracy equations, compared to the SOTA autoregressive models, and such capability is crucial for SR.

	\end{itemize}

\section{Related work}

\subsection{Mathematical Expression Synthesis}

Latent variable models provide a probabilistic framework for uncovering hidden structure in symbolic data.
The CVAE~\cite{gomez2018automatic} encodes equations represented as strings into a continuous latent space with sequence model and then decodes them back into discrete equations, optimizing via the Evidence Lower Bound. However, string-based representations often struggle to guarantee syntactic correctness. To address this, several works~\cite{dai2018syntax,kusner2017grammar} leverage context-free grammars by representing equations as sequences of grammar production rules, enforcing syntactic validity. The hierarchical VAE~\cite{mevznar2023efficient} explicitly models expressions by encoding their inherent tree-structured representations.

Autoregressive-based language models have expanded to symbolic calculations, like symbolic integration~\cite{DBLP:conf/iclr/LampleC20}, and multi-step arithmetic reasoning~\cite{trinh2024solving}.
The hierarchy of expression is embedded in the generation process. 
Hms~\cite{lin2021hms} mitigates error propagation by organizing semantic understanding into three cascading layers to engage its dependency-aware tree decoder.
NER-HRT~\cite{zhang2024number} enhances this pipeline by incorporating explicit numerical embeddings and a two-stage recursive decoding approach, achieving improved accuracy.

\subsection{Deep Symbolic Regression}
Deep learning methods for SR mainly include reinforcement learning~(RL) and supervised learning methods. 
DSR~\cite{petersen2019deep} trains an autoregressive RNN to emit a vector of logits, using a risk-seeking policy gradient method.
NGGP~\cite{mundhenk2021symbolic} initializes the GP population with samples from the RL agent and uses the best GP individuals to train the network further.
DisCo-DSO~\cite{pettit2025disco} proposes joint discrete-continuous optimization to determine both skeletons and constants through RL training.
To filter the meaningless information from high-noise data, NRSR~\cite{sun2025noise} leverages a noise resilient gating module in RL training. 
Large-scale pre-training models for SR have been proposed.
Conditioned on the numerical data points, autoregressive Transformers have achieved significant generalization performance.
Based on the feature of numerical data processed by the encoder,
the decoder in NeSymReS~\cite{biggio2021neural} autoregressively predicts the token of mathematical skeleton symbols. 
E2E~\cite{kamienny2022end} goes beyond predicting
operators; it also predicts the exact values of constants.
Recent work has utilized contrastive learning to exploit cross-modal representations~\cite{meidani2023snip}, bridging numerical simulations and symbolic equations.
To further improve the accuracy of the pre-trained model, TPSR~\cite{shojaee2023transformer} accelerates the MCTS process with the pre-trained Transformer model.

\subsection{Diffusion Model for Text Generation}
Diffusion models \cite{ho2020denoising}, which learn to invert a diffusion process that gradually corrupts data into Gaussian noise, have demonstrated remarkable performance in continuous domains such as natural image generation~\cite{saharia2022photorealistic,wang2025stay}. Extending diffusion models to discrete domains like text generation has prompted various exploratory approaches. The primary methodologies include discrete diffusion and continuous diffusion techniques. In discrete diffusion, data is corrupted by switching from one discrete value to another through transition kernels~\cite{nie2025large,austin2021structured}. Alternatively, continuous diffusion treats the embeddings of discrete texts, applying the diffusion process directly to these representations~\cite{li2022diffusion,gong2022diffuseq}. In this work, we adopt the latter approach, applying continuous diffusion to the embeddings of mathematical tokens.

\section{Preliminary}
\subsection{Task Description} Symbolic regression~(SR) is integral to understanding how variables mathematically interact within a physical system, which forms a fundamental part of the scientific method.
Given a set of $N$ data points, where each point has features $Z_i \in \mathbb{R}^D$ and target $y_i\in\mathbb{R}$, symbolic regression aims to find the explicit mapping mathematical equation $f_i$ that maps these features to their respective targets: $y_i\approx f_i(Z_i)$, for all points $i\in N$. 

\subsection{Diffusion Model}
In this paper, we leverage the denoising
diffusion probabilistic model (DDPM)~\cite{ho2020denoising} to model the probability distribution of mathematical equations. The diffusion process can be regarded as a discrete-time Markov process. In the forward process, $x_0$ is diffused in a $T$ time steps  Markov
chain by gradually interpolating the iterate with Gaussian noise, following a pre-defined sequence of noise levels, $\beta_1,...\beta_T$.
Specifically, at each time step $t$, the transition of  latent variables $x_{t-1} \rightarrow x_{t}$ is parametrized by:

\begin{equation}
\label{forward_eq}
q(\vx_{t} |\,\vx_{t-1}) = \mathcal{N}(\sqrt{1-\beta_t} \vx_{t-1}, \beta_t\bm{I}).
\end{equation}
Due to the Markov property and the Gaussian transitions, we can 
 obtain the $\vx_t$ at any step $t$ as a function of $\vx_0$: $q(\vx_t | \vx_0) = \mathcal{N}(\vx_t; \sqrt{\bar{\alpha}_t} \vx_0, (1 - \bar{\alpha}_t) \bm{I})$,
where $\bar{\alpha}_t=\prod_{i=1}^t \alpha_i$, $\alpha_t= 1 - \beta_t$.

 The reverse process is defined as approximately inverting the diffusion process in Eq.~\ref{forward_eq}. 
Starting from $\vx_T \sim \mathcal{N}(0, \bm{I})$, the reverse process aims to recover the initial data $\vx_0$ through denoising:

\begin{equation} 
\begin{split}
p_\theta(\vx_{0:T}):=p(\vx_T)\prod_{t=1}^Tp_{\theta}(\vx_{t-1}|\vx_t),
p_\theta(\vx_{t-1} | \vx_t) \\
= \mathcal{N}(\vx_{t-1}; \boldsymbol{\mu}_\theta(\vx_t, t), \sigma_t^2 \bm{I}) 
\end{split}
\end{equation}
where $\boldsymbol{\mu}_\theta(\vx_t, t)$ and $\boldsymbol{\sigma}_t^2$ 
is the predicted parameterization of the mean and standard variation of $q(\vx_{t} |\,\vx_{t-1})$ in the forward process.
In this work, the learning of $p_\theta$ is based on a Transformer architecture $f_\theta$ to estimate $\vx_0$ in every step, $f_\theta(\vx_t, t)$.  The denoising transition from $\vx_{t}$ to 
 $\vx_{t-1}$ can be formulated as:

\begin{equation}
\label{reverse_eq}
\begin{split}
p_\theta(\vx_{t-1}|\vx_t)=\mathcal{N}(\mathbf{x}_{t-1};\frac{\sqrt{\overline\alpha_{t-1}}\beta_t}{1-\overline\alpha_t}f_\theta(\vx_t,t) \\
+ \frac{\sqrt{\alpha_t}(1-\overline\alpha_{t-1})}{1-\overline\alpha_t}\vx_t,\frac{1-\overline\alpha_{t-1}}{1-\overline\alpha_t}\beta_t)
\end{split}
\end{equation}

As shown in~\cite{ho2020denoising}, training can be simplified by minimizing the mean squared error  between the ground truth data $\vx_0$ and its estimates: $\mathbb{E}_{\vx_0 \sim q(\vx_0),\, t \sim \mathcal{U}(1, T)}\|\vx_0-f_\theta(\vx_t, t)\|^2\cdot$

\section{Method}

\begin{figure*}[h]
\centering		\includegraphics[scale=0.4]{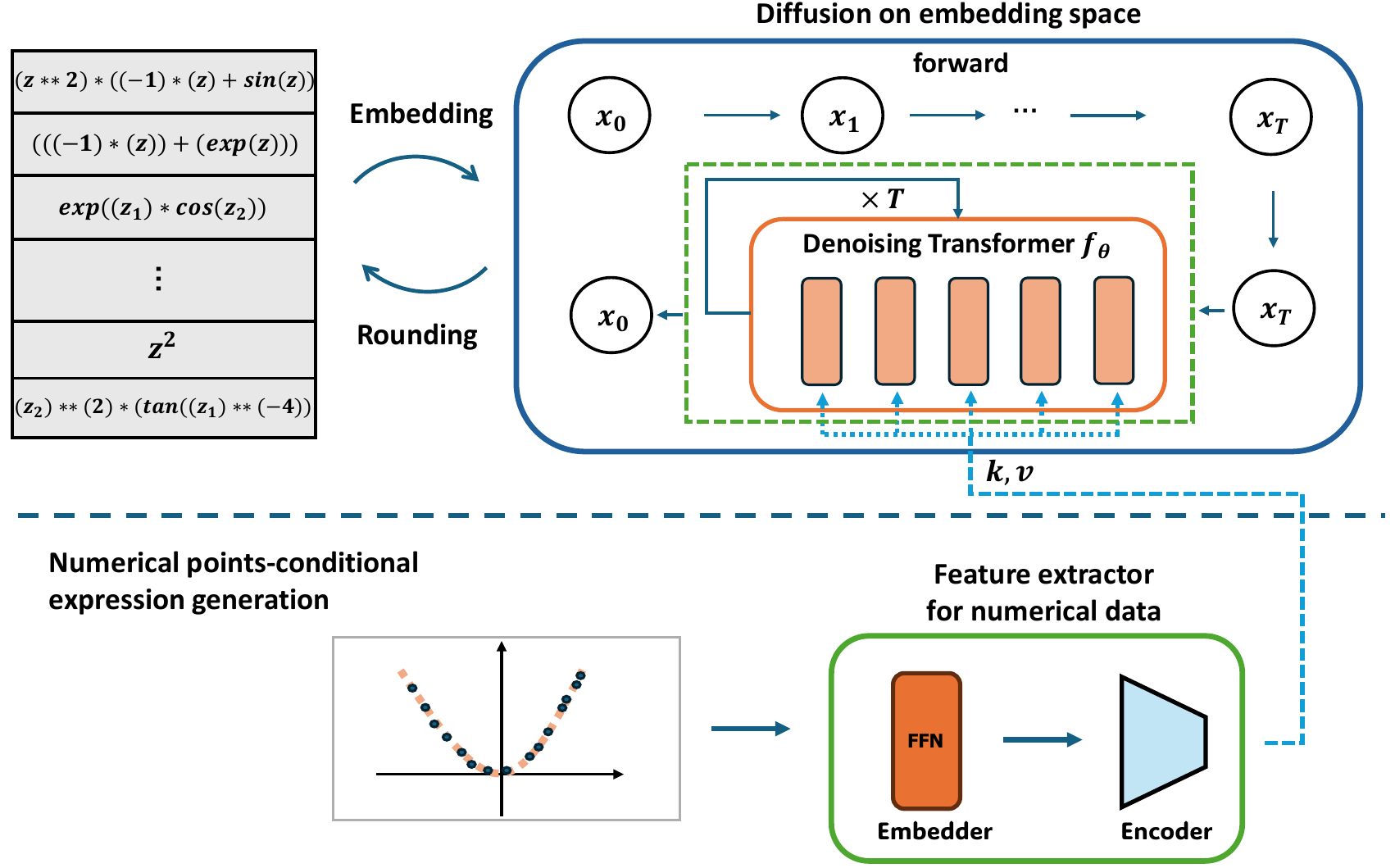}

\caption{
  Overview of the DiffuSR framework. The schematic above the dashed line is the process for unconditional expression generation. 
  Below the dashed line is the numerical data point feature processing, which provides generation guidance for SR.
  }
\label{method}
\end{figure*}

Fig.~\ref{method} illustrates the architecture of our method. We use a trainable embedding layer to map the discrete mathematical tokens into continuous latent space, following the prominent work on diffusion for text~\cite{li2022diffusion,gong2022diffuseq}.
The forward Markov chain in diffusion progressively destroys the latent embedding by injecting noise, and the reverse process learns to reconstruct these embeddings.
We leverage the diffusion language model to solve a numeric-to-symbolic equation generation task, SR, which is a goal-directed equation synthesis task conditioned on numerical points. The feature of data points is leveraged as guidance to generate equations to fit the numerical data.

\subsection{Data Creation and Tokenization}
The mathematical equation data are generated by randomly sampling mathematical operators as internal nodes and constants or variables as leaves. Details are provided in the Experimental Setup section.
Equations are represented in pre-fix order sequences and the discrete mathematical symbols with continuous numerical constants are tokenized following the vocabulary in~\cite{kamienny2022end}, where every mathematical operator and variable takes one token and a constant occupies three tokens with its sign, mantissa (between 0 and 9999), and
exponent (from E-100 to E100).




\subsection{Embedding and Rounding}
After tokenization, the equation sequences $\mathbf{w}=[w_1,...,w_L]$ are treated
as a sequence of words, where $L$ denotes the max length of the equation. Each token $w_i$ has an associated
embedding $\ve_{i} \in \mathbb{R}^d $ with dimension $d$, transformed by a trainable function: $Emb(w_i)=\ve_i$.
We define the diffusion process in the embedding space.
To adapt the discrete mathematical equations input into the standard continuous diffusion process, the starting point for the forward process emerges by sampling from a Gaussian distribution centered on the embedding matrix $\mathbf{E}=[\ve_{1},...,\ve_{N}], \in \mathbb{R}^{N\times D}$: $x_0 \sim \mathcal{N}(\mathbf{E},\sigma_0\bm{I})$.

To map vectors in embedding space to discrete words, a rounding step is introduced using a projection onto the vocabulary. This is achieved through a linear layer followed by a softmax function:

\begin{equation} \label{eq
} p(\mathbf{w} | \vx_0) = Softmax(W^\top \vx_0), \end{equation}
where $W$ is the weight matrix, shared with the embedding function $Emb$ to ensure consistency between the embedding space and the output space.





\subsection{Numerical Data Points Guidance}
To incorporate numerical data guidance to
shape the equation generation process, we first transform the numerical data points into a latent space. We leverage a frozen encoder~\cite {kamienny2022end} to extract features from the data points.
To mitigate the quadratic complexity of Transformers as $D$ and $N$ increase, each input point is mapped to a single embedding. The $N$ pairs of numerical data points $\{Z_i, y_i\}_N \in \mathbb{R}^{D+1}$, represented as $3\times(D+1)$ tokens, are first processed by a 2-layer fully connected feedforward network. The resulting $N$ embeddings are then fed to a 4-layer Transformer encoder, to obtain the final numerical data features $\mathbf{N} \in \mathbb{R}^{N \times d}$, where $d=512$ is the dimension of embedding. To maintain the permutation invariance of data points,  the positional embeddings are removed from the encoder.

\textbf{Model.} We leverage a 12-layer Transformer with cross-attention to make the equation generator attend to the numerical data features $\mathbf{N}$. 
The diffusion step $t$ will be mapped to  
time embedding $TE$, and the sequence of input equation tokens encoded to positional embedding $PE$. During each denoising step, the transform of current diffusion state $\vx_t$ combined with $TE$ and $PE$ will be treated as queries: $\mathbf{Q}=W_Q\cdot[W_{in}(\vx_t)+TE+PE]$. 
The conditioning embeddings $\mathbf{N}$ provide the keys $\mathbf{K}=W_K\cdot MLP(\mathbf{N})$ and values $\mathbf{V}=W_V \cdot MLP(\mathbf{N}$).
The $W_{in}$, $W_Q$, $W_K$, $W_V$ are learnable projection matrices. The cross-attention mechanism is computed using the following formula:

\begin{equation}
\text{Attention}(\mathbf{Q},\mathbf{K},\mathbf{V})=softmax(\frac{\mathbf{Q}\mathbf{K}^T}{\sqrt{d_k}})\mathbf{V}
\end{equation}



\textbf{Training.}
The training objective is to minimize
the squared error between the predicted and true $\vx_0$, as
well as the reconstruction error between $f_{\theta}(\vx_1, 1; \mathbf{N})$ and the target
embeddings $\mathbf{E}$. Building on  \cite{ho2020denoising} and \cite{li2022diffusion}, the loss function can be written as:

\begin{equation}
\begin{split}
\mathcal{L}=\mathop{\mathbb{E}}_{q(\vx_{0:T}|\mathbf{w})} \Big[\sum_{t=2}^T||\vx_0-f_{\theta}(\vx_t, t; \mathbf{N})||^2 +  \\
||\mathbf{E}-f_{\theta}(\vx_1, 1; \mathbf{N})|| {^2} 
- \log p_{\theta}(\mathbf{w}|\vx_0)\Big]
\end{split}
\end{equation}

\subsection{Decoding Strategies}
To obtain the symbolic equation from the trained
model, we start from the final step $T$ and sample a
state $\vx_T$ from a standard Gaussian distribution. Then the learned reverse transition, as defined in Eq.~\ref{reverse_eq}, will be applied iteratively to recover $\vx_0$. 
The resulting mathematical equation will be obtained by decoding the emitted logits matrix $p(\mathbf{w} | \vx_0)$ greedily. 
However, the results obtained from the greedy rounding process might fail to achieve high accuracy, due to the sample diversity of the probabilistic generative model.
Therefore, specialized decoding strategies are designed for SR.

\textbf{Top-K Sampling.} We run the pre‑trained diffusion model $\mathcal{K}$ times, each run starting with an independent Gaussian noise vector $\mathbf{x}_T^{(k)}$ determined by a different random seed. 
Within every run, we follow the model’s greedy choice at each denoising step, which yields $\mathcal{K}$ candidate equations $\{\mathcal{E}^{(k)}\}_{k=1}^{\mathcal{K}}$. 
The decoded equations consist of operators, variables, integers, and floating-point constants.
Following the prior work~\cite{kamienny2022end}, the predicted constants will be initialized to the Broyden-Fletcher-Goldfarb-Shanno (BFGS) algorithm for further refinement. The candidate equation with the best accuracy will be reported.


\textbf{Diffusion Guided GP Search.}
To further refine the solution obtained by greedy decoding, we introduce GP to explore the operator space more thoroughly.
Classic GP searches from scratch and is data-inefficient; we aim to bootstrap the search with predictions from our trained diffusion model.
We seed the population with the diffusion’s best decoded equation into a dedicated subpopulation with the size of $\mathcal{L}$, following~\cite{han2025transformer}.
During evolution, we replace GP’s uniform subtree mutation with a neural network guided variant: at the randomly chosen mutation node, we sample the replacement root from the logits distribution $p(\mathbf{w} | \vx_0)$, using masks that keep only tokens allowed by the grammar. Children are generated recursively from the same logits until a height limit is reached, guaranteeing every offspring is syntactically valid. The pseudocode of this decoding method is shown in the Appendix.



%

\section{Experiments}
This section reports the empirical evaluation of DiffuSR. 
We first verify its unconditional generation quality-how well the model captures the marginal distribution of mathematical equations, without numeric data conditioning.
Subsequently, we evaluate DiffuSR on standard SR benchmarks.


\subsection{Experimental Setup}\label{Experimental setup}
\textbf{Training Data Generation.}
Following prior works~\cite{DBLP:conf/iclr/LampleC20,bendinelli2023controllable}, we first sample expression skeleton trees with operators as internal nodes and variables $\{z_d\}_{d \leqslant 3}$ or the constant placeholder \texttt{`c'} as leaves.
The maximum equation length is 20, and the max appearances of non-leaf nodes is 5.
Internal nodes are sampled from an unnormalized weighted distribution, listed in Table~\ref{Operators}, and the leaves are sampled from variables and the constant placeholder. Prefix equations like 
\texttt{\seqsplit{[`mul', `c', `pow', `x\_1', `2']}}
are created via pre-order traversal of the skeleton trees and further simplified using the \textit{simplify} function in Sympy\footnote{\url{https://www.sympy.org/}}.
The placeholder \texttt{`c'} is then replaced with a concrete float number by sampling from $\mathcal{U}(-10,10)$ for additive constants, while multiplicative constants are sampled using a logarithmic distribution $\mathcal{U}(0.05,10)$.
For SR, the training dataset also includes numerical data, $\{Z_i, y_i\}_N$, where $y_i=f_i(Z_i)$.
The numerical data points are obtained by sampling each independent variable from a
uniform distribution $\mathcal{U}(-10,10)$ in $N$ runs, where $N$ is a random positive integer not exceeding 1000.
Then the dependent variable $y_i$ is computed by evaluating the equation $f_i$ using the sampled inputs $Z_i$.
We generate \textit{10 million} random training equations for both unconditional generation and SR. 


\begin{table}[htbp]
    \centering
    \small
    {\setlength{\tabcolsep}{15pt}
    \begin{tabular}{c|c}
    
        \toprule
			
			\textbf{Arity} & \textbf{Operators  }\\
			
			\hline
			Binary & \texttt{mul:10, div:5, add:10, sub:5}  \\ 
             \hline
			Unaray & 
 
 \texttt{\makecell{exp:4,sin:4,cos:4,tan:4, asin:2,  \\ sqrt:4, pow2:5, pow3:2, ln:1} }
 
 \\ 
         
        \bottomrule
    \end{tabular}
    }
       \caption{Operators with un-normalized sampling probability}
    \label{Operators}
\end{table}

\textbf{Method Settings.}
For the training model, we use a Transformer that consists of 12 hidden layers, each equipped with 12 attention heads, totaling 118M parameters.
The embedding size $d$ for the diffusion process is 128, projected to a hidden size of 768, with an intermediate size of 3072. We utilize a square root noise schedule~\cite{li2022diffusion} with 2000 diffusion steps in training.
The model employs the GELU activation function, dropout rates of 0.1 on attention and hidden layers to mitigate overfitting,  and layer normalization with an epsilon value of 1e-12.
We initialize weights with a standard deviation of 0.02 and apply an exponential moving average for model updates at a rate of 0.9999. Training uses the AdamW optimizer with a learning rate of 5e-5 and a batch size of 128.

The decoding parameters for SR are set as follows. $\mathcal{K}$ is set to 20 in Top-K sampling. For Diffusion Guided GP Search, the greedy solution from the diffusion model's inference is replicated $\mathcal{L} = 10$ times.
During mutation, each mutated node has a $50\%$ probability of undergoing a diffusion-guided mutation; otherwise, a random mutation is applied. This process is executed in parallel across 10 independent instances with different seeds using the Ray library.\footnote {\url{https://github.com/ray-project/ray}.}


\subsection{Unconditional Equation Synthesis}
 Can our method model the marginal distribution over mathematical equations?
This establishes a syntactic and structural basis.
Experiments are conducted to measure the quality of sampled equations.
The generated equations must adhere to grammatical correctness, as this serves as a foundation. Further, the equations are desired to
exhibit diversity. We define diversity in equations from two perspectives: Complexity and Syntax, and employ the following metrics.

$\bullet$ {\bf Grammar Valid Rate}. 
This metric assesses the percentage of generated equations that are syntactically valid according to established mathematical grammar rules. It ensures that the models produce structurally correct formulas, serving as a baseline for further evaluations.

$\bullet$ {\bf Diversity in Syntax}. 
To evaluate the diversity of syntax, we employ the Self-BLEU metric~\cite{papineni2002bleu}, originally proposed for testing diversity in text generation, to measure symbolic-level diversity among equations. 

$\bullet$ {\bf Diversity in Complexity}.
The complexity of equations is evaluated based on the number of nodes in the expression tree generated~\cite{la2021contemporary}. We aim for the generated results to encompass formulas of diverse complexities. To measure this, we calculate the entropy of the complexities of the equations:
$-\sum c(x) \log c(x)$, where $c(x)$ denotes the probability of encountering a formula with a particular complexity. 


\textbf{Results.}
We compare our method with existing equation generators, Character VAE~\cite{gomez2018automatic} and VAE for hierarchical data~(HVAE)~\cite{mevznar2023efficient}. 
For each generative method, 1000 equations are sampled, with results summarized in Table~\ref{unconditional_result}. The `type' refers to whether the training data consists of basic skeleton equations or equations with fully specified constants. 
Notably, our method achieves high grammatical validity rates and demonstrates higher diversity than the compared methods, for both skeleton and equation with precise constants generation.
The baseline methods were designed solely to generate equation skeletons. 
CVAE performs poorly in terms of grammatical validity.
Benefiting from learning the tree-structured data, HVAE excels in generating valid formula skeletons. While it only supports generating skeletons and exhibits lower diversity. The example generations and further analysis of our model are presented in the Appendix.

\begin{table}[h]

\small
\centering
\def\arraystretch{1.1}
\begin{tabular}{c|c|c|c|c }
 
    \toprule
    \textbf{Methods} & \textbf{Type} & \textbf{Valid(\%) $\uparrow$} & \textbf{\makecell{Self-\\BLEU }$\downarrow$} & \textbf{\makecell{Complexity\\ entropy}$\uparrow$}  \\
    \hline

CVAE & Skele & 19.8 & 0.432
 & 2.51 \\ \hline
    
HVAE & Skele & 100 & 0.86
 & 2.83 \\ \hline
 
\multirow{2}{*}{\makecell{Ours}} & Skele & 100 & 1.11e-233 & 2.87  \\ \cline{2-5}
    & Full & 99.9 & 1.07e-232 & 2.88 \\
    \bottomrule
\end{tabular}
\caption{ 
The percentages of valid equations and data on diversity indicators of 1000 sampled equations.}
\label{unconditional_result}
\end{table}

\begin{table*}[!h]
\small
\centering
\def\arraystretch{1.2}

    \begin{tabular}{l|l|ll|ll|ll|ll|ll}
       \toprule[1.5pt]
       \multirow{2}{*}{Method} & \multirow{2}{*}{Type}  & \multicolumn{2}{c}{Nguyen} & \multicolumn{2}{c}{Livermore} & \multicolumn{2}{c}{Constant}  & \multicolumn{2}{c}{Jin} & \multicolumn{2}{c}{Avg} \\
         &  & $R^2$ & $\mathcal{C}$  & $R^2$ & $\mathcal{C}$  & $R^2$ & $\mathcal{C}$  & $R^2$ & $\mathcal{C}$  & $R^2$ & $\mathcal{C}$    \\
        \hline
        GP  & Heuristic & {0.893} & {36.43}  & {0.755} & {44.31}  & {0.818} & {34.29}  & {0.763} & {154.10}  & {0.810} & {67.29}  \\
          \hline

\makecell{SymGPT}& AR Transformer* + S & {0.609} & {28.60}  & {0.536} & {49.88}  & {0.713} & {42.5}  & {0.576} & {39.85} & {0.609} & {40.21}  \\
NeSym  & AR Transformer* + BS & {0.975} & {11.92}  & {0.941} & {12.22}  & {0.967} & {16.88}  & {0.786} & {15.68} & {0.917} & {14.18}  \\
E2E  & AR Transformer* + S & {0.911} & {26.82}  & {0.837} & {31.39}  & {0.964} & {23.78}  & {0.779} & {26.47} & {0.873} & {27.12}  \\
\hline
HVAE  & VAE*+ ES & {0.989} & {15.85}  & {0.893} & {16.91}  & {0.984} & {16.13}  & {0.833} & {19.64} & {0.925} & {17.13}  \\
        \hline
DSR  & RL + G& {0.954} & {15.68}  & {0.904} & {17.52}  & {0.955} & {16.25}  & {0.803} & {18.75} & {0.904} & {17.05}  \\
NGGP  & RL + GP & {0.979} & {17.50}  & {0.910} & {23.82}  & {0.989} & {18.25}  & {0.875} & {16.83} & {0.938} & {19.10} \\
\hline
TPSR  & AR Transformer* + MCTS & {0.989} & {32.24}  & {0.948} & {38.36}  & \textbf{0.999} & {32.14}  & {0.957} & {47.93}   & {0.973} & {37.67}\\
\hline
DiffuSR\textsuperscript{1}  & Diffusion* + G & {0.986} & \textbf{8.58}  & {0.943} & \textbf{9.41}  & {0.964} & \textbf{10.63}  & {0.797} & \textbf{14.5} & {0.923} & \textbf{10.61}  \\
DiffuSR\textsuperscript{2}  &Diffusion* + GP & \textbf{0.998} & {16.66}  & \textbf{0.995} & {20.51}  & \textbf{0.999} & {19.08}  & \textbf{0.973} & {28.23} & \textbf{0.992} & {21.12} \\

        \bottomrule[1.5pt]
    \end{tabular}
    \caption{Results of DiffuSR with two decoding ways and existing baselines.
    * represents the pretrained model, and the other methods are learning or searching from scratch.
    RL, G, S, BS, ES, and MCTS denote Reinforcement Learning, Greedy decoding, Sampling Search, Beam-Search, Evolutionary Search, and Monte Carlo Tree Search, respectively.
    DiffuSR\textsuperscript{1} and DiffuSR\textsuperscript{2} represent decoding the pre-trained diffusion model with top-20 sampling and diffusion-guided GP search, respectively.
    }
   \label{benchmark_res}
\end{table*}

\subsection{Symbolic Regression Experiments}\label{conditional}

In this section, we evaluate various methods on the goal-directed equation synthesis task, SR.
The objective of SR is to discover mathematical equations with both high accuracy and high interpretability. 
We assess the performance of DiffuSR on a series of well-known symbolic
regression benchmark suites: Nguyen~\cite{uy2011semantically},  Livemore~\cite{mundhenk2021symbolic}, Constant~\cite{mundhenk2021symbolic}, and Jin~\cite{jin2019bayesian}, totaling 48 datasets.
Each dataset contains a ground-truth equation and its corresponding sampled data points, which are split into a training set and a test set, as detailed in the Appendix.

\textbf{Metrics.} 
To measure accuracy, we utilize the coefficient of determination ($R^2$), which quantifies the proportion of variance in the actual values explained by predictions. $R^2$ serves as an indicator of model fit: $R^2 = 1 - \frac{\sum_{i=1}^{N_{test}}{(y_i - \hat{y}_i)}^2}{\sum_{i=1}^{N_{test}}{(y_i - \bar{y})}^2} \nonumber$, 
where the $N_{test}$ is the number of test data pairs, $y_i$ refers to the actual values, $\hat{y}_i$ denotes the predicted values, and the $\bar{y}$ is the mean of the actual values.
To assess interpretability, we employ the complexity metric~\cite{la2021contemporary}, $\mathcal{C}$, defined as the number of nodes in the generated expression tree. The complexity is usually seen to be inversely proportional to the interpretability of an equation.

\textbf{Competing Methods.}
We compare our model with several representative methods for SR. Three representative autoregressive Transformer models are considered,   
\textbf{SymbolicGPT}~\cite{valipour2021symbolicgpt}, \textbf{NeSymReS}~\cite{biggio2021neural} and \textbf{E2E}~\cite{kamienny2022end}.  
Combining the autoregressive Transformer model with external search, \textbf{TPSR}~\cite{shojaee2023transformer} leverages a pre-trained model to enhance the Monte Carlo Tree Search.
\textbf{GP}~\cite{koza1994genetic} represents programs as syntax trees and utilizes genetic operators to evolve new programs. The GP method is implemented by \textit{GPlearn}\footnote{\url{https://gplearn.readthedocs.io/}.} framework, and the evolutionary parameters are set as it is recommended. 
\textbf{DSR}~\cite{petersen2019deep} is a representative RL-based method for SR, and \textbf{NGGP}  ~\cite{mundhenk2021symbolic} further combines RL with GP to improve accuracy.
Based on the pre-trained model for unconditional equation synthesis, the \textbf{HVAE}~\cite{mevznar2023efficient} is applied to the SR task, using the evolutionary algorithm to search the latent space and the decoder to obtain equations. Since the network predicts only the formula skeleton, HVAE additionally evolves constant values within each generation. All testing methods share the instructions: \{$+$, $-$, $\times$, $\div$, $sin$, $cos$, $tan$, $exp$, $log$, $sqrt$, $asin$, $acos$, $atan$, $abs$, $pow$\}.




 %
 %
    




\textbf{Benchmark Results.}
The average $R^2$ and $\mathcal{C}$ values for all methods on the test datasets are listed in Table~\ref{benchmark_res}. 
Among all methods, DiffuSR with guided GP search consistently achieves the highest fitting accuracy. In particular, it surpasses other hybrid approaches that also combine learning models with symbolic search, including NGGP and TPSR, in both accuracy and complexity.
DiffuSR with top-20 sampling achieves higher accuracy compared to other end-to-end pretrained models. Notably, it outperforms all baseline methods in terms of complexity, as indicated by much lower values of $\mathcal{C}$. This reduction in complexity is crucial for practical applications, as shorter equations tend to be more interpretable.
The $R^2$ gap between the two decoding strategies highlights the effectiveness of diffusion-guided GP search, despite the increased complexity introduced by randomness.

\begin{figure}[h]
		\centering   
\subfigure[]{\includegraphics[width=0.45\linewidth]{ 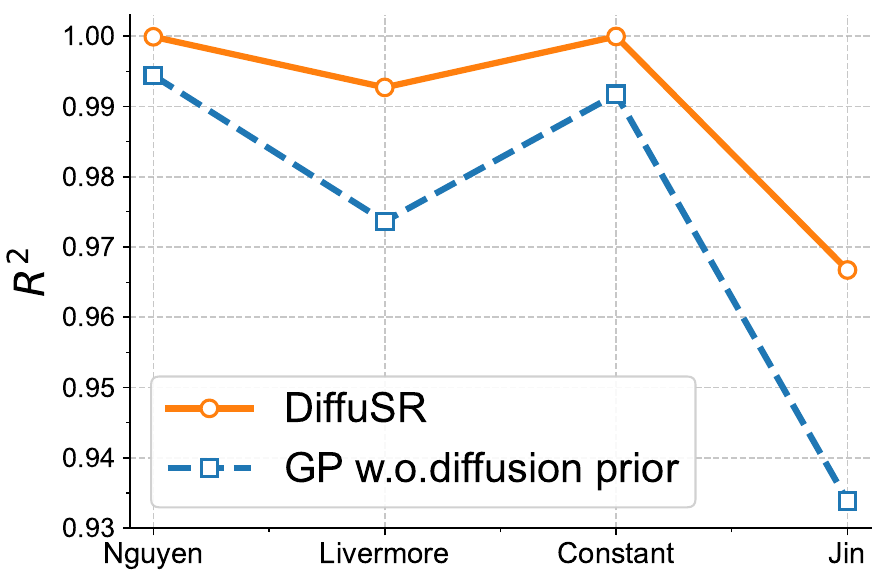}}
\subfigure[]
  {\includegraphics[width=0.46\linewidth]{ 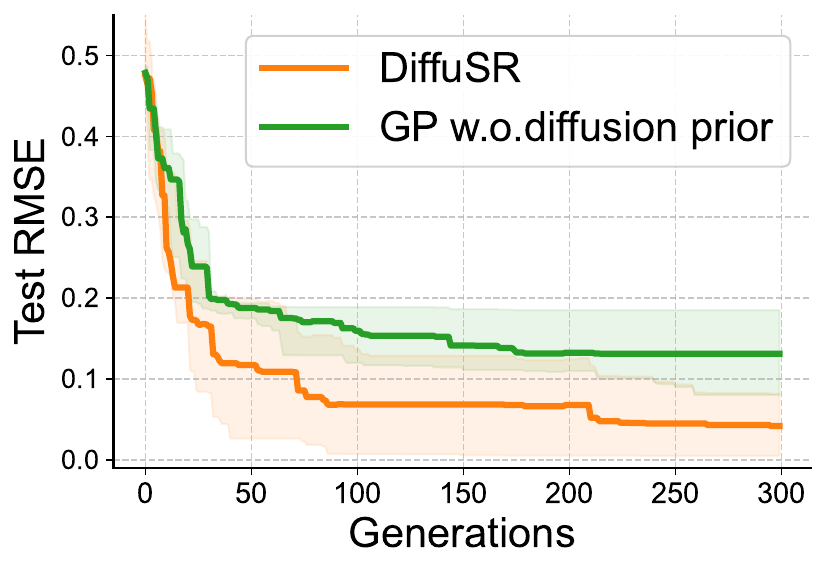}}

		\caption{Mean $R^2$ scores and evolutionary $RMSE$ of DiffuSR using guided GP search versus random GP mutation. }
        \label{ablation_r2}
  \label{sinhx}
	\end{figure}

\begin{table*}[!t]
\centering
\small
\def\arraystretch{1.1}

{\setlength{\tabcolsep}{0pt}
\begin{tabularx}{\textwidth}{
  >{\centering\arraybackslash}m{0.18\textwidth}|
  >{\centering\arraybackslash}m{0.41\textwidth}|
  >{\centering\arraybackslash}m{0.41\textwidth}
}
\toprule[1.5pt]

\textbf{Models} & $\mathbf{sinh(x)}$ & $\mathbf{cosh(x)}$ \\
\midrule

\textbf{NeSymReS} & $x \cdot \exp(0.16246 \cdot x^2)$ & $\exp(0.509387\cdot x \cdot \sin(x))$ \\
\hline
\multirow{2}{*}{\centering\makecell{\textbf{E2E}\\ (simplified)}}& \begin{tabular}[c]{@{}c@{}} $0.000448*x^4 + 0.159328*x^3$ \\ $+ 0.005227*x^2 + 1.012409*x$ \\ $+ 0.0015267$ \end{tabular} & \begin{tabular}[c]{@{}c@{}} $0.5391\cdot x^2$ \\ $+ 0.0017251\cdot x$ \\$+ 1$ \end{tabular} \\
\hline
\multirow{3}{*}{\textbf{DiffuSR}} & $x + 0.17336 \cdot x^3$ & $x^2 + \cos(x)$ \\
\cline{2-3}  
& $x \cdot \exp(0.16246 \cdot x^2)$ & $x^2 + \cos(\sin(x))$ \\
\cline{2-3} 
& $x^2 \cdot \sin(x)^{-1}$ & $x^2 \cdot \sin(x)^{-2}$ \\
\bottomrule[1.5pt]
\end{tabularx}

\caption{
Predicted equations with a fitting accuracy of $R^2 > 0.99$ from large-scale pre-trained models on the \textit{Livermore-7} and \textit{Livermore-8} datasets. The functions discovered by DiffuSR are listed in order of their $R^2$ values.
}
\label{diversity_case}
}

\end{table*}
\textbf{Ablation for Diffusion Guided Mutation.}
To study the effect of the pre-trained model in the diffusion-guided GP search, ablation experiments are conducted: we replace the sampling nodes from the logits distribution
$p(\mathbf{w} | \vx_0)$ with a random subtree mutation in classic GP.
The mean $R^2$ curve and evolutionary root mean square error plots for the two methods are shown in Fig.~\ref {ablation_r2}.
We find that the network-guided mutation enables GP to converge to a higher accuracy level more quickly than an unguided search that has to discover promising structures purely at random. This biased sampling shrinks the search space and raises the average quality of newly generated individuals.

\begin{figure}[!h]
  \centering
  \subfigure[]{\includegraphics[scale=0.23]{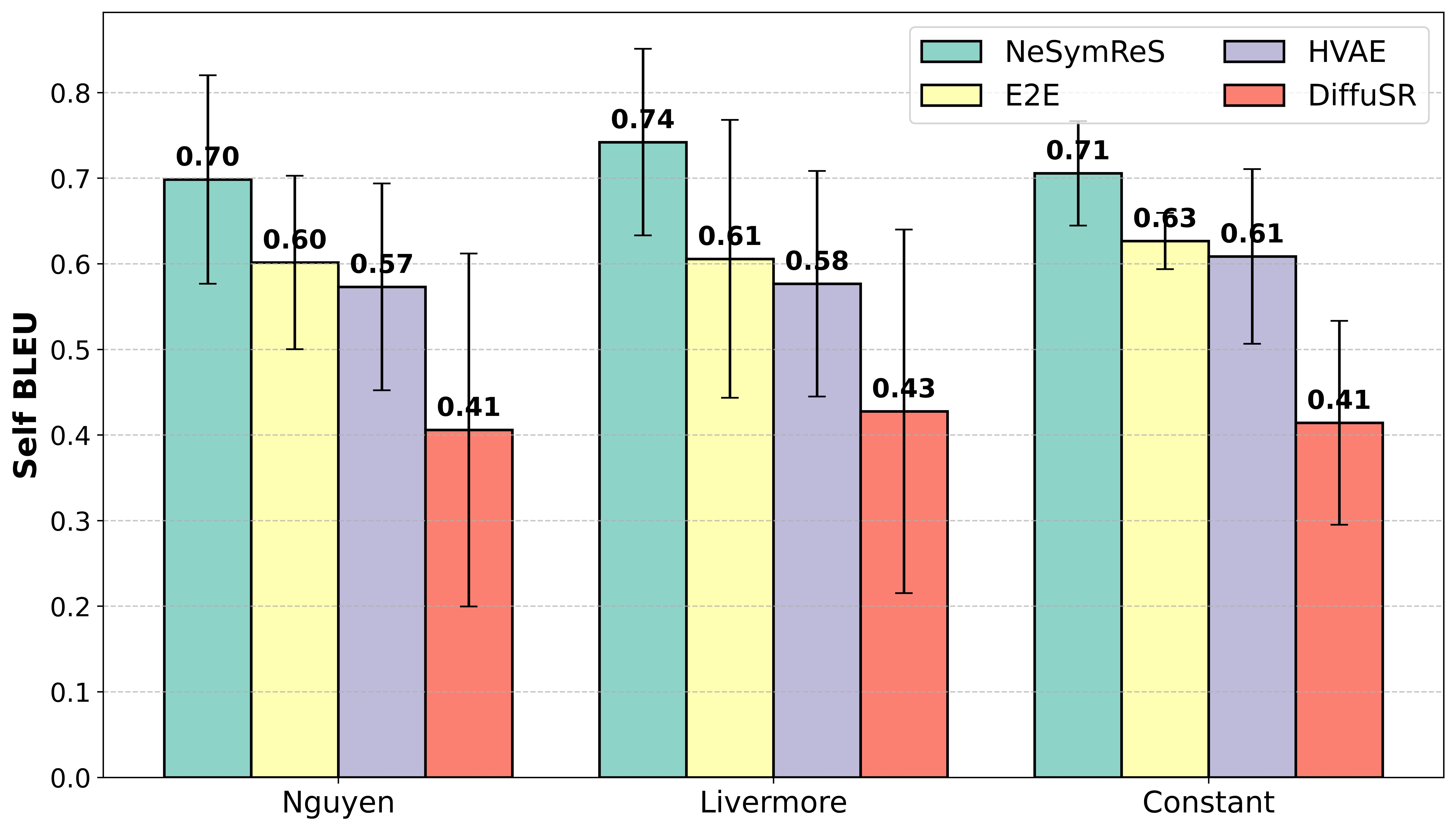}} 
  \subfigure[]{\includegraphics[scale=0.3]{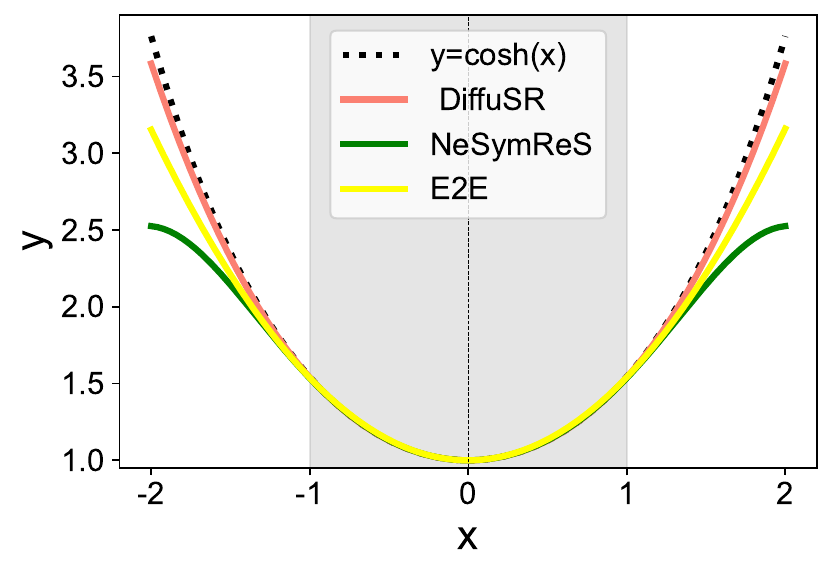}
  \includegraphics[scale=0.3]{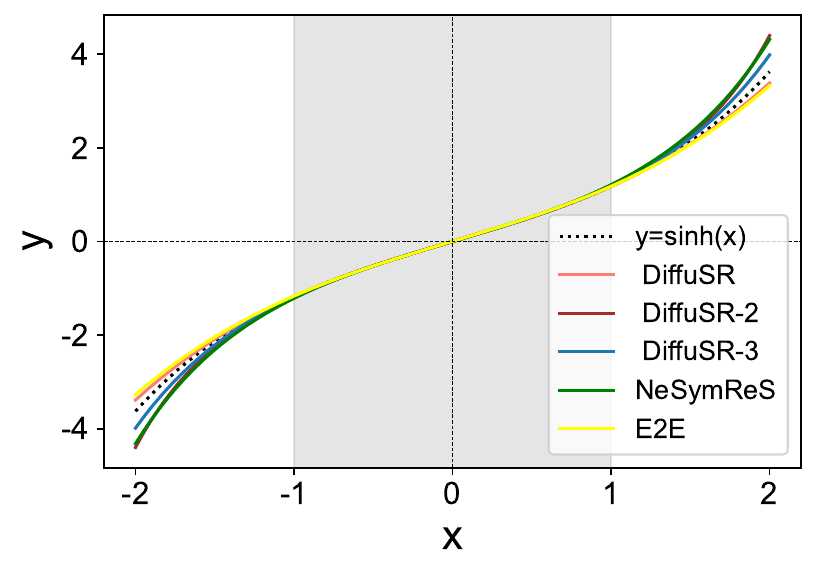}
  }
  

  \caption{ (a) 
  Mean Self-BLEU for pre-trained SR models across benchmark series, with error bars denoting the standard deviations. (b) Extrapolation performance comparison with models and the black-box truth functions $cosh(x)$ and $sinh(x)$. The shaded region denotes the range of the test dataset, [-1, 1], which comprises 200 data points. }
 \label{SR-diversity}

\end{figure}

\textbf{Finding Equivalent yet Diverse Forms.}
We compared the Self-BLEU scores of the top 10 equations generated by HVAE, autoregressive models, and DiffuSR with greedy decoding.
In Fig.~\ref{SR-diversity}(a), DiffuSR exhibits the lowest Self-BLEU values, indicating greater syntactic diversity in its predictions.
To further explore the advantages of diversity, we present qualitative studies in Table~\ref{diversity_case}, which lists the top predictions with high fitting accuracy ($R^2 > 0.99$) from end-to-end SR models.
We observe that autoregressive models tend to converge to similar equations across runs, whereas DiffuSR is capable of generating a wider variety of equations that are both diverse and highly accurate.
As shown in Fig.~\ref{SR-diversity}(b), this diversity enhances the model's ability to recover and understand the underlying black-box formula.

In the shown cases, it is challenging to find the true equation,
because the symbol operators, $sinh$ and $cosh$, are not included in the vocabulary of all models.
Therefore, the goal shifts to finding mathematically equivalent equations with different forms.
Although all models achieve high $R^2$ values, only DiffuSR successfully produces nearly equivalent approximations.
As shown  in Table~\ref{diversity_case}, for $sinh(x)$, the best prediction by DiffuSR: $ x + 0.17336 \cdot x^3 $ serves a third-degree polynomial that approximates the Taylor expansion $sinh(x)\approx x+\frac{x^3}{6} $.
 The predicted coefficient $0.17336$ is close to $\frac{1}{6} \approx 0.16667$, demonstrating a meaningful approximation.
 Similarly, for $cosh(x)$, the prediction from DiffuSR, $x^2+cos(x)$ approximates $cosh(x)\approx 1+\frac{x^2}{2}$ by substituting the $cos(x)$ with its Taylor expansion $cos(x)\approx 1-\frac{x^2}{2}$.
In contrast, autoregressive models fail to learn the approximations; their predictions diverge from the true functions outside the domain of the test data, as shown in Fig.~\ref{SR-diversity}(b).

\section{Conclusion}
In this paper, we present DiffuSR, a novel diffusion model for symbolic regression.
We embed mathematical equation sequences and optimize embeddings based on numerical datapoints.
Based on the large-scale pretraining, DiffuSR exhibits good generalization to unseen SR tasks and outperforms the autoregressive model in terms of complexity and diversity.
We also design a decoding strategy to combine the strengths of the pre-trained model and the search ability of GP.
The results show that DiffuSR can find diverse, concise, and high-accuracy expressions, compared to the SOTA
autoregressive models. 
Yet, due to the challenges of SR, 
multiple inference attempts are often required to obtain high-accuracy equations, and improving the model's stability and sampling speed remains an important goal in future work.


\appendix
\bibliography{mybibliography}

\begin{thebibliography}{38}
\providecommand{\natexlab}[1]{#1}

\bibitem[{Austin et~al.(2021)}]{austin2021structured}
Austin, J.; et~al. 2021.
\newblock Structured denoising diffusion models in discrete state-spaces.
\newblock \emph{Advances in neural information processing systems}, 34: 17981--17993.

\bibitem[{Bendinelli et~al.(2023)}]{bendinelli2023controllable}
Bendinelli, T.; et~al. 2023.
\newblock Controllable neural symbolic regression.
\newblock In \emph{International Conference on Machine Learning}, 2063--2077. PMLR.

\bibitem[{Biggio et~al.(2021)}]{biggio2021neural}
Biggio, L.; et~al. 2021.
\newblock Neural symbolic regression that scales.
\newblock In \emph{International Conference on Machine Learning}, 936--945. PMLR.

\bibitem[{Chudnovsky et~al.(2020)}]{chudnovsky2020computer}
Chudnovsky, D.~V.; et~al. 2020.
\newblock Computer algebra in the service of mathematical physics and number theory.
\newblock In \emph{Computers in mathematics}, 109--232. CRC Press.

\bibitem[{Dai et~al.(2018)}]{dai2018syntax}
Dai, H.; et~al. 2018.
\newblock Syntax-directed variational autoencoder for structured data.
\newblock \emph{arXiv preprint arXiv:1802.08786}.

\bibitem[{G{\'o}mez-Bombarelli et~al.(2018)}]{gomez2018automatic}
G{\'o}mez-Bombarelli, R.; et~al. 2018.
\newblock Automatic chemical design using a data-driven continuous representation of molecules.
\newblock \emph{ACS central science}, 4(2): 268--276.

\bibitem[{Gong et~al.(2024)}]{gong2024text}
Gong, H.; et~al. 2024.
\newblock Text-guided molecule generation with diffusion language model.
\newblock In \emph{Proceedings of the AAAI Conference on Artificial Intelligence}, volume~38, 109--117.

\bibitem[{Gong et~al.(2022)}]{gong2022diffuseq}
Gong, S.; et~al. 2022.
\newblock Diffuseq: Sequence to sequence text generation with diffusion models.
\newblock \emph{arXiv preprint arXiv:2210.08933}.

\bibitem[{Han et~al.(2025)}]{han2025transformer}
Han, X.; et~al. 2025.
\newblock Transformer-Assisted Genetic Programming for Symbolic Regression [Research Frontier].
\newblock \emph{IEEE Computational Intelligence Magazine}, 20(2): 58--79.

\bibitem[{Ho et~al.(2020)}]{ho2020denoising}
Ho, J.; et~al. 2020.
\newblock Denoising diffusion probabilistic models.
\newblock \emph{Advances in Neural Information Processing Systems}, 33: 6840--6851.

\bibitem[{Jin et~al.(2019)}]{jin2019bayesian}
Jin, Y.; et~al. 2019.
\newblock Bayesian symbolic regression.
\newblock \emph{arXiv preprint arXiv:1910.08892}.

\bibitem[{Kahlmeyer et~al.(2025)}]{kahlmeyer2025dimension}
Kahlmeyer, P.; et~al. 2025.
\newblock Dimension Reduction for Symbolic Regression.
\newblock In \emph{Proceedings of the AAAI Conference on Artificial Intelligence}, volume~39, 17707--17714.

\bibitem[{Kamienny et~al.(2022)}]{kamienny2022end}
Kamienny, P.-A.; et~al. 2022.
\newblock End-to-end symbolic regression with transformers.
\newblock \emph{Advances in Neural Information Processing Systems}, 35: 10269--10281.

\bibitem[{Koza(1994)}]{koza1994genetic}
Koza, J.~R. 1994.
\newblock Genetic programming as a means for programming computers by natural selection.
\newblock \emph{Statistics and computing}, 4: 87--112.

\bibitem[{Kusner et~al.(2017)}]{kusner2017grammar}
Kusner, M.~J.; et~al. 2017.
\newblock Grammar variational autoencoder.
\newblock In \emph{International Conference on Machine Learning}, 1945--1954. PMLR.

\bibitem[{La~Cava(2023)}]{la2023flexible}
La~Cava, a.~o. 2023.
\newblock A flexible symbolic regression method for constructing interpretable clinical prediction models.
\newblock \emph{NPJ Digital Medicine}, 6(1): 107.

\bibitem[{La~Cava et~al.(2021)}]{la2021contemporary}
La~Cava, W.; et~al. 2021.
\newblock Contemporary symbolic regression methods and their relative performance.
\newblock \emph{Advances in Neural Information Processing Systems}, 2021: 1.

\bibitem[{Lample et~al.(2020)}]{DBLP:conf/iclr/LampleC20}
Lample, G.; et~al. 2020.
\newblock Deep Learning For Symbolic Mathematics.
\newblock In \emph{International Conference on Learning Representations, April 26-30}.

\bibitem[{Li et~al.(2022)}]{li2022diffusion}
Li, X.; et~al. 2022.
\newblock Diffusion-lm improves controllable text generation.
\newblock \emph{Advances in Neural Information Processing Systems}, 35: 4328--4343.

\bibitem[{Lin et~al.(2021)Lin, Huang, Zhao, Chen, Liu, Wang, and Wang}]{lin2021hms}
Lin, X.; Huang, Z.; Zhao, H.; Chen, E.; Liu, Q.; Wang, H.; and Wang, S. 2021.
\newblock Hms: A hierarchical solver with dependency-enhanced understanding for math word problem.
\newblock In \emph{Proceedings of the AAAI conference on artificial intelligence}, volume~35, 4232--4240.

\bibitem[{Meidani et~al.(2023)}]{meidani2023snip}
Meidani, K.; et~al. 2023.
\newblock Snip: Bridging mathematical symbolic and numeric realms with unified pre-training.
\newblock \emph{arXiv preprint arXiv:2310.02227}.

\bibitem[{Me{\v{z}}nar et~al.(2023)}]{mevznar2023efficient}
Me{\v{z}}nar, S.; et~al. 2023.
\newblock Efficient generator of mathematical expressions for symbolic regression.
\newblock \emph{Machine Learning}, 112(11): 4563--4596.

\bibitem[{Mundhenk et~al.(2021)}]{mundhenk2021symbolic}
Mundhenk, T.~N.; et~al. 2021.
\newblock Symbolic regression via neural-guided genetic programming population seeding.
\newblock \emph{arXiv preprint arXiv:2111.00053}.

\bibitem[{Nie et~al.(2025)}]{nie2025large}
Nie, S.; et~al. 2025.
\newblock Large Language Diffusion Models.
\newblock \emph{arXiv preprint arXiv:2502.09992}.

\bibitem[{Papineni et~al.(2002)}]{papineni2002bleu}
Papineni, K.; et~al. 2002.
\newblock Bleu: a method for automatic evaluation of machine translation.
\newblock In \emph{Proceedings of the 40th annual meeting of the Association for Computational Linguistics}, 311--318.

\bibitem[{Petersen et~al.(2019)}]{petersen2019deep}
Petersen, B.~K.; et~al. 2019.
\newblock Deep symbolic regression: Recovering mathematical expressions from data via risk-seeking policy gradients.
\newblock \emph{arXiv preprint arXiv:1912.04871}.

\bibitem[{Pettit et~al.(2025)}]{pettit2025disco}
Pettit, J.~F.; et~al. 2025.
\newblock Disco-dso: Coupling discrete and continuous optimization for efficient generative design in hybrid spaces.
\newblock In \emph{Proceedings of the AAAI Conference on Artificial Intelligence}, volume~39, 27117--27125.

\bibitem[{Reddy et~al.(2025)}]{reddy2025towards}
Reddy, C.~K.; et~al. 2025.
\newblock Towards scientific discovery with generative ai: Progress, opportunities, and challenges.
\newblock In \emph{Proceedings of the AAAI Conference on Artificial Intelligence}, volume~39, 28601--28609.

\bibitem[{Saharia et~al.(2022)}]{saharia2022photorealistic}
Saharia, C.; et~al. 2022.
\newblock Photorealistic text-to-image diffusion models with deep language understanding.
\newblock \emph{Advances in Neural Information Processing Systems}, 35: 36479--36494.

\bibitem[{Shojaee et~al.(2023)}]{shojaee2023transformer}
Shojaee, P.; et~al. 2023.
\newblock Transformer-based planning for symbolic regression.
\newblock \emph{Advances in Neural Information Processing Systems}, 36: 45907--45919.

\bibitem[{Sun et~al.(2025)}]{sun2025noise}
Sun, C.; et~al. 2025.
\newblock Noise-Resilient Symbolic Regression with Dynamic Gating Reinforcement Learning.
\newblock In \emph{Proceedings of the AAAI Conference on Artificial Intelligence}, volume~39, 20690--20698.

\bibitem[{Trinh et~al.(2024)}]{trinh2024solving}
Trinh, T.~H.; et~al. 2024.
\newblock Solving olympiad geometry without human demonstrations.
\newblock \emph{Nature}, 625(7995): 476--482.

\bibitem[{Udrescu et~al.(2020)}]{udrescu2020ai}
Udrescu, S.-M.; et~al. 2020.
\newblock AI Feynman 2.0: Pareto-optimal symbolic regression exploiting graph modularity.
\newblock \emph{Advances in Neural Information Processing Systems}, 33: 4860--4871.

\bibitem[{Uy et~al.(2011)}]{uy2011semantically}
Uy, N.~Q.; et~al. 2011.
\newblock Semantically-based crossover in genetic programming: application to real-valued symbolic regression.
\newblock \emph{Genetic Programming and Evolvable Machines}, 12: 91--119.

\bibitem[{Valipour et~al.(2021)}]{valipour2021symbolicgpt}
Valipour, M.; et~al. 2021.
\newblock Symbolicgpt: A generative transformer model for symbolic regression.
\newblock \emph{arXiv preprint arXiv:2106.14131}.

\bibitem[{Virgolin et~al.(2022)}]{virgolin2022symbolic}
Virgolin, M.; et~al. 2022.
\newblock Symbolic regression is NP-hard.
\newblock \emph{arXiv preprint arXiv:2207.01018}.

\bibitem[{Wang et~al.(2025)}]{wang2025stay}
Wang, R.; et~al. 2025.
\newblock STAY Diffusion: Styled Layout Diffusion Model for Diverse Layout-to-Image Generation.
\newblock In \emph{2025 IEEE/CVF Winter Conference on Applications of Computer Vision (WACV)}, 3855--3865. IEEE.

\bibitem[{Zhang et~al.(2024)}]{zhang2024number}
Zhang, Y.; et~al. 2024.
\newblock Number-enhanced representation with hierarchical recursive tree decoding for math word problem solving.
\newblock \emph{Information Processing \& Management}, 61(2): 103585.

\end{thebibliography}

\section{Appendix}

\section{A. Details of Diffusion Guided GP Search}
We present the pseudocodes of the Diffusion Guided GP Search method, as shown in Algorithm.~\ref{DGGP}

\begin{algorithm}[htbp]
		\caption{Diffusion Guided GP Search}\label{DGGP}
		\renewcommand{\algorithmicrequire}{\textbf{Input:}}
		\renewcommand{\algorithmicensure}{\textbf{Output:}}
		\begin{algorithmic}[1]
			\REQUIRE Logits matrix $p(\mathbf{w} | \vx_0)$, Rate $\delta$, Height limitation $H$, Crossover rate $r_c$, Mutation rate $r_m$, Generations $N$
			\STATE \textit{G}   $\leftarrow$ \textit{greedy decode} ($p(\mathbf{w} | \vx_0)$)
            \STATE \textit{P}  $\leftarrow$  $\{ clone(G) \}_{\mathcal{L}} \cup $
            $\{$ Random Trees $\}_{P-\mathcal{L}}$
            
            \FOR{\textit{gen} $= 1,..., N$ }
            \STATE {\textcolor{gray}{\slshape \# selection}}
            \STATE \textit{parents} $\leftarrow$ \textit{tournament selection (P)}
            
            \STATE {\textcolor{gray}{\slshape \# crossover}}
                \IF {$rand() \leq r_c$}
                \STATE \textit{P} $\leftarrow$ \textit{P}  $\cup$  \textit{crossover (parents)}
			\ENDIF
                \STATE {\textcolor{gray}{\slshape \# mutation}}
                \FOR{$I \in parents$}
                \IF {$rand() \leq r_m$}
                
			\IF {$rand() \leq \delta$}
                \STATE {\textcolor{gray}{\slshape \# determine the mutation position and height}}
			\STATE \textit{pos} $\leftarrow rand\_int$~$(1,|$I$|)$
            
                \STATE $h_t$ $\leftarrow  rand\_int(1,H)$
            
                \STATE {\textcolor{gray}{\slshape \# subtree growth guided by the diffusion}}
                \STATE \textit{new subtree} $\leftarrow$ GROW~( $p(\mathbf{w} | \vx_0)$ , \textit{pos} , $h_t$  )
                \STATE {\textcolor{gray}{\slshape \# replace the original subtree with new tree
                }}

			\STATE $I_{new}$ $ \leftarrow replace\_subtree(n,\textit{new subtree})$
			
            \ELSE
        
			\STATE $I_{new}$ $\leftarrow$ \textit{random mutation} $(I)$
			\ENDIF

			\STATE $I$  $\leftarrow$  $I_{new}$ 
			
			\ENDIF 
            \ENDFOR

        \STATE \textit{P} $\leftarrow$ \textit{elitism(P)}
			
\ENDFOR
      \RETURN best sample      
		\end{algorithmic}
	\end{algorithm}

\begin{algorithm}[h]
		\caption{\textit{Function} GROW }
		\renewcommand{\algorithmicrequire}{\textbf{Input:}}
		\renewcommand{\algorithmicensure}{\textbf{Output:}}
		\begin{algorithmic}[1]
			\REQUIRE Logits matrix $p(\mathbf{w} | \vx_0)$, Mutation position $n$, Height $h$, Unary operators set $\mathbb{U}$, Binary operators set $\mathbb{B}$, Leaf nodes set $\mathbb{L}$
            \STATE subtree $\leftarrow$ []
            \STATE {\textcolor{gray}{\slshape \# determine the type of new node
            }}
            \IF {$h\leq 0$}
            \STATE \textit{category mask} $\leftarrow$ \textit{Leaf}
            \ELSE
            \STATE \textit{category mask} $\leftarrow$ \textit{Operator}
            \ENDIF

        \STATE {\textcolor{gray}{\slshape \# sample a node according to the $n$ row of logits matrix}}
        \STATE \textit{token} $\leftarrow$ \textit{sample} from $\mathbb{B} \cup \mathbb{U} \cup \mathbb{L} $ according to: 
        \\
        \hspace{3.5em} SOFTMAX$(p(\mathbf{w} | \vx_0)_n *$ \textit{category mask)};
        \STATE subtree.append(\textit{child})

        \IF { \textit{token} $\in \mathbb{U}$ }

        \STATE \textit{child} $\leftarrow$ GROW~($n+1$, $h-1$)
        \STATE subtree.append(\textit{child})

        \ELSIF{\textit{token} $\in \mathbb{B}$}
        \STATE \textit{left} $\leftarrow$ GROW~($n+1$, $h-1$)

        \STATE \textit{right} $\leftarrow$ GROW~($n+1+len(left)$, $h-1$)

        \STATE subtree.append(\textit{left,} \textit{right})

        \ENDIF

        \RETURN subtree
		\end{algorithmic}
	\end{algorithm}

\section{B. Impact of Numerical Data Encoder}
Two types of pre-trained encoders for extracting features of numeric data are explored: the encoders from
E2E\footnote {\url{https://github.com/facebookresearch/symbolicregression}.} and SNIP\footnote{\url{https://github.com/deep-symbolic-mathematics/Multimodal-Symbolic-Regression}.}.
E2E is trained solely for SR using a cross-entropy loss and comprises a numerical data encoder and decoder to generate expression.  
In contrast, SNIP includes a numerical data encoder and a symbolic encoder, aiming to maximize the similarity between matching numeric data-symbolic expression pairs while minimizing it for non-matching pairs in training.
We trained DiffuSR using these encoders separately, and the training plots are shown in Figure.~\ref{encoders}.
It is evident that DiffuSR with the E2E encoder achieves a higher $R^2$ score than with the SNIP encoder. Besides, the SNIP encoder tends to exhibit instability during training.
The SR experiments in SNIP are conducted by exploring the latent space combined with gradient-free optimization methods, making it challenging to assess the impact of its numerical data encoder on the end-to-end SR task. 
However, it is clear that the choice of encoder significantly influences the expression generation task, paralleling findings in text-to-image diffusion models.
\setcounter{figure}{3} 
\begin{figure}[h]
  \centering
  \subfigure[]{\includegraphics[scale=0.16]{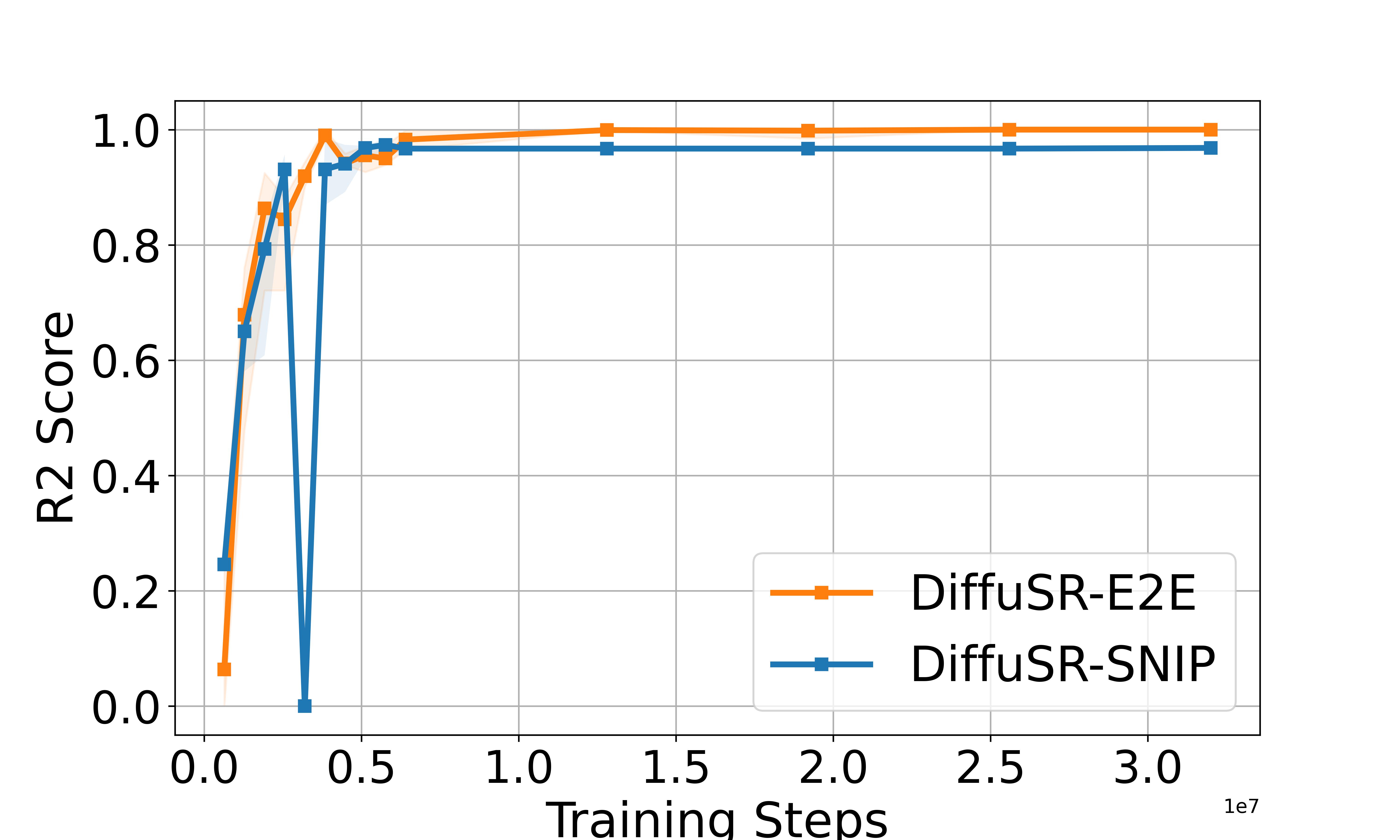}} 
  \subfigure[]{\includegraphics[scale=0.16]{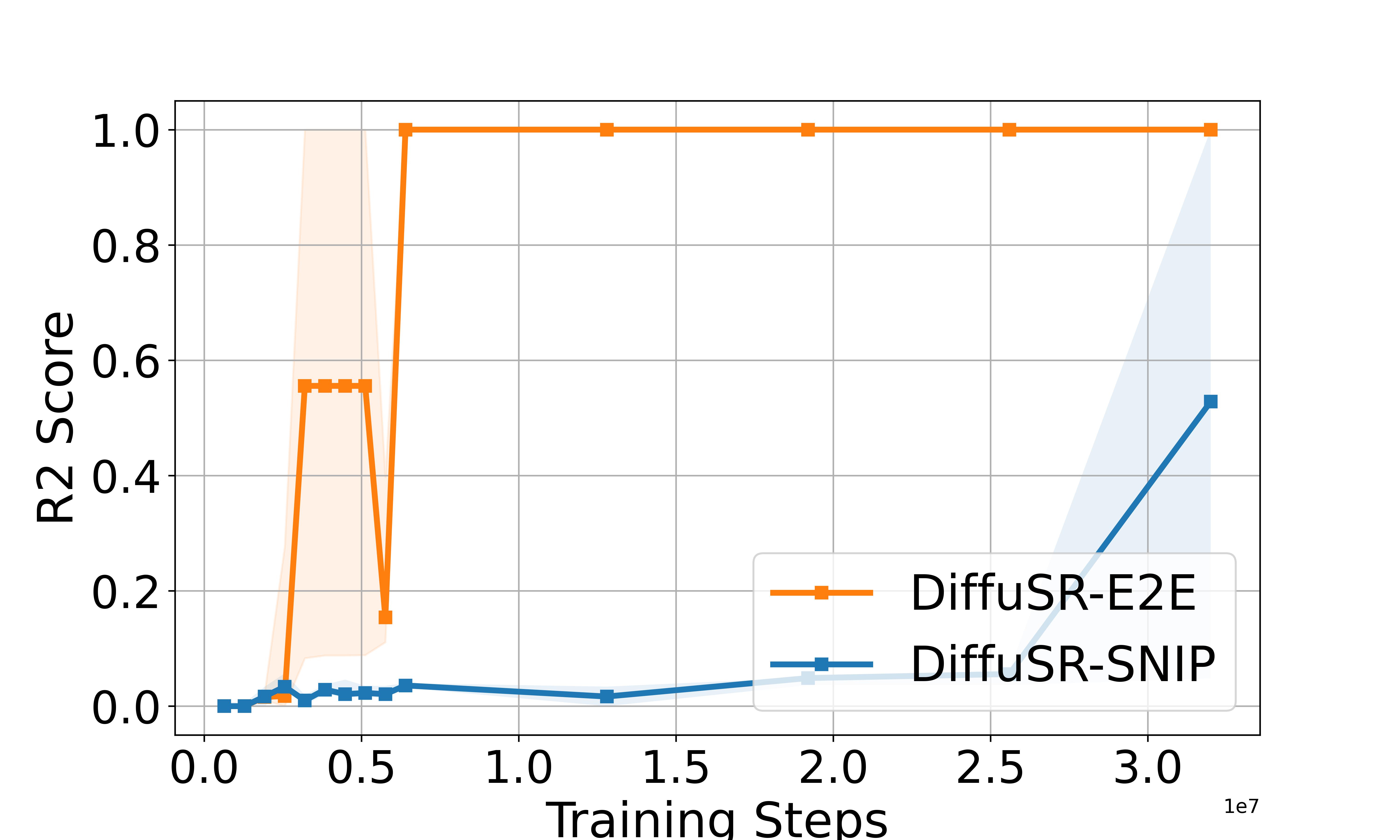}}  
  \caption{Comparison between numerical data encoders for SR. DiffuSR models with the E2E and SNIP encoders during training are tested on \textit{Nguyen-1}, and \textit{Livermore-11} datasets. Median $R^2$ scores are plotted, with the [25\%, 75\%] percentile intervals shaded.}
 \label{encoders}

\end{figure}

\setcounter{table}{3} 
\begin{table*}[!h]
\centering

\label{generation_exapmles}
\renewcommand\arraystretch{1.5}
\begin{tabularx}{\textwidth}{ >{\hsize=0.3\hsize\centering\arraybackslash}X | >{\hsize=0.7\hsize\centering\arraybackslash}X }
\toprule
\textbf{Method}  & \textbf{Generated Expressions}\\
\midrule
CVAE & $x_2^{((c+x_1)+3)}$\\ \hline
HVAE& $(x_1+x_2)+(c*x_1)$\\ \hline
Ours-skele  & $sin(x_1^2)*(x_2)^{-1}$\\ \hline
Ours-full  & $(x_3+(x_2^2)^{-1})*(-3.454+x_1+(-1)*sin(x_2))$\\ 
\bottomrule
\end{tabularx}
\caption{Examples of generated mathematical expressions from different methods.}
\end{table*}

\section{C. Additional Analysis for Unconditional Expression Synthesis}
\begin{table*}[!h]

\centering

\renewcommand{\arraystretch}{1.5}
\begin{tabularx}{\textwidth}{
  >{\hsize=0.12\hsize\centering\arraybackslash}X|
  >{\hsize=0.44\hsize\centering\arraybackslash}X|
  >{\hsize=0.44\hsize\centering\arraybackslash}X
}

\toprule
\textbf{Step} & \textbf{Ours-skele} & \textbf{Ours-full} \\
\midrule
$T=500$ & \texttt{[\textcolor{red}{`mul'}, \textcolor{red}{`x\_1'}, \textcolor{red}{`mul'}, \textcolor{red}{`x\_2'}, \textcolor{red}{`x\_2'}, \textcolor{red}{`add'}, \textcolor{red}{`c'}, \textcolor{red}{`x\_3'}]}  & \texttt{[`mul', `+', \textcolor{red}{`mul'}, `E-3', `pow', `add', `x\_2', `add', `mul', `+', \textcolor{red}{`mul'}, `E-4', `x\_1', `mul', `+', \textcolor{red}{`mul'}, `E-4', `pow', `x\_1', `2']} \\
$T=1000$ & \texttt{ [`mul', `x\_1', `add', `x\_2', `mul', `-1', `exp', `x\_2']}
\newline \newline \textit{Infix:} $ x_1 \cdot (x_2 + (-1) \cdot \exp(x_2)) $ 
& \texttt{[`mul', `+', `N1134', `E-3', `pow', `add', `-1', `add', `mul', `+', \textcolor{red}{`x\_1'}, `E-4', `x\_1', `mul', `+', \textcolor{red}{`mul'}, `E-5', `mul', `x\_1', `x\_2', `-2']}\\
$T=1500$ & \texttt{ [`mul', `x\_1', `add', `1', `mul', `-1', `exp', `x\_2']} \newline \newline \textit{Infix:} $ x_1 \cdot (1 + (-1) \cdot \exp(x_2)) $ & \texttt{[`mul', `+', 'N1134', `E-3', `pow', `add', `x\_2', `add', `mul', `+', `N2141', `E-4', `x\_1', `mul', `+', \textcolor{red}{`add'}, `E-5', `mul', `x\_1', `x\_2', `-2']}\\ 

$T=2000$ & \texttt{[`mul', `x\_1', `add', `1', `mul', `-1', `exp', `x\_2']} \newline \newline \textit{Infix:} $ x_1 \cdot (1 + (-1) \cdot \exp(x_2)) $& \texttt{[`mul', `+', `N2700', `E-3', `pow', `add', `x\_2', `add', `mul', `+', `N2042', `E-4', `x\_1', `mul', `+', `N2421', `E-5', `mul', `x\_1', `x\_2', `-2']}
\newline \newline \textit{Infix:} $ 2.7 \cdot (x_2 + (0.2042 \cdot x_1 + 0.02421 \cdot x_1 \cdot x_2^{-2}))  $
\\
\bottomrule
\end{tabularx}
\caption{The generated expressions from different diffusion steps. Marked in red is the invalid expression. We further report the infix form of the generated valid expression.}
\label{denoising_words}
\end{table*}

We investigate how the DiffuSR performs under different numbers of inverse diffusion steps.
An example of the predicted initial diffusion embeddings, rounded to symbol words at denoising steps of 500, 1000, 1500, and 2000, from both models is listed in Table~\ref{denoising_words}.
Clearly, fewer diffusion steps significantly degrade the quality of generated equations, especially for DiffuSR-full.
Moreover, generating expressions with full constants is more challenging than generating skeletons only.
As illustrated in Table~\ref{denoising_words}, DiffuSR-skele produces grammatically valid equations within  1000 steps, while DiffuSR-full encounters errors in predicting constants, even at higher steps.
Fig.~\ref{denoise_plot} presents the median grammatical validity rate of 100 generated expressions for both models across multiple runs at various diffusion steps. Consistent with earlier observations, DiffuSR-skele achieves a higher validity rate with fewer steps, whereas DiffuSR-full requires more.

Two factors may contribute to this phenomenon.
Firstly, the unique encoding mechanism for constants necessitates three consecutive legal tokens to represent a float, such as $0.2042$ being expressed as [`+’, `N2042’, `E-4’], which makes the generation more challenging.
Secondly, the vocabulary for fully specified constants is considerably larger than that for skeleton formulas. The `skele' methods utilize a shared vocabulary of 35 tokens, including operators, integers between -10 and 10, and the constant placeholder \texttt{`c'}. In contrast, the `full' methods utilize a vocabulary that includes over 10,000 additional tokens representing constants, expanding the complexity of the model's output space.
\setcounter{figure}{4} 
\begin{figure}[!h]
		\centering
		\includegraphics[scale=0.3]{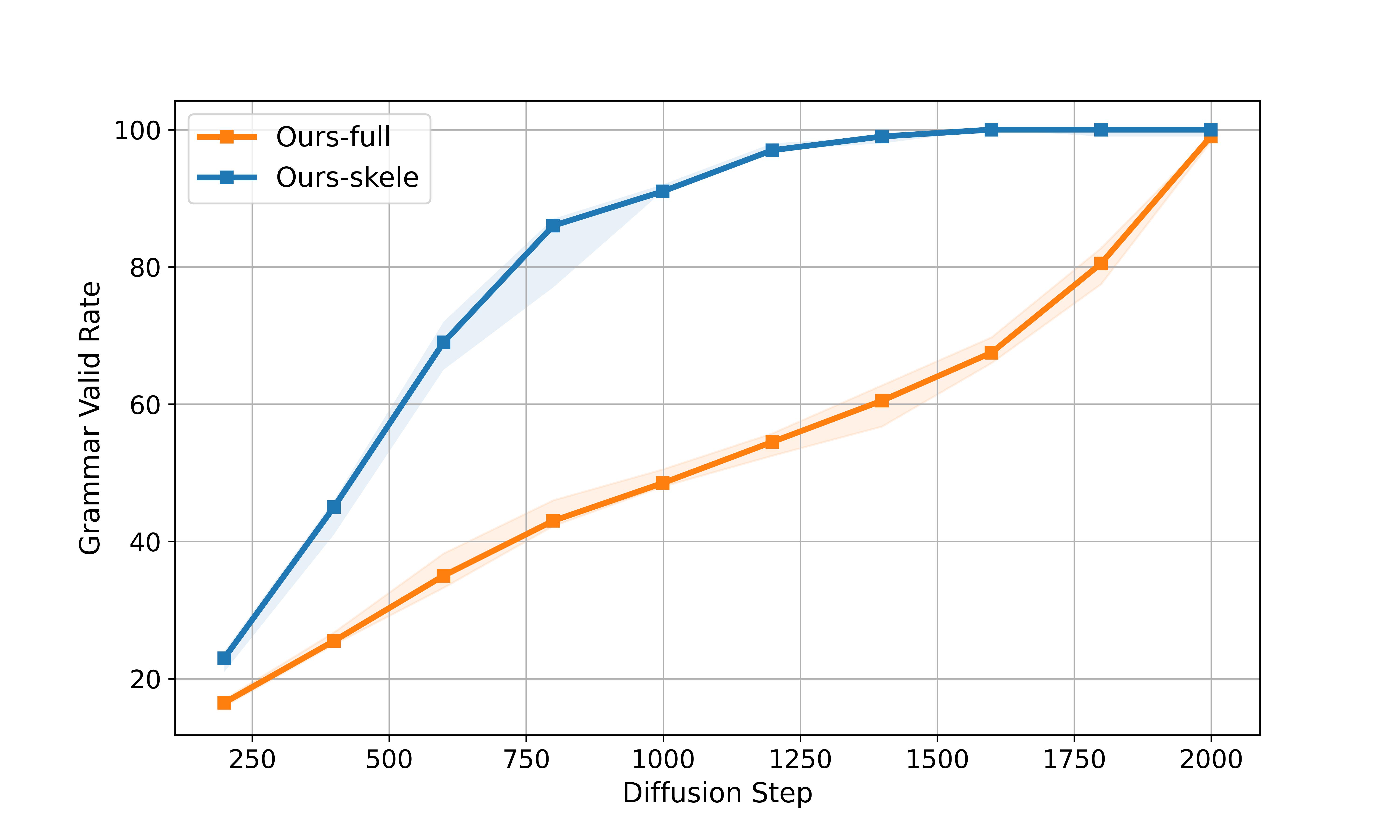}
       
		\caption{Median grammar valid rate of 100 generated expressions from different diffusion steps,  [25\%,75\%] percentile interval is shaded.}
  \label{denoise_plot}
	\end{figure}

\section{D.  Details of Genetic Programming Parameters}
\begin{table}[!h]
\renewcommand\arraystretch{1.2}
		\centering
        \small
		
        \resizebox*{0.85\linewidth}{!}{
		\begin{tabular}{c|c}
			\hline
			\textbf{Parameter} & {\textbf{Value}} \\ \hline
			Population Size &  300 \\ \hline
			Crossover Rate  &  0.5~      \\ \hline
			Mutation Rate &  0.5 \\
             \hline
			Generations   &     300   \\ \hline
			Max program height    &    7   \\ \hline
			Initialization method  &    \makecell{Ramped Half-and-Half, \\height range 2-6}  \\ \hline
			Selection method     &  \makecell{ Tournament selection, \\tournament size 3}    \\ \hline
			Fitness Function      &  RMSE      \\ \hline

		\end{tabular}\label{GP_settings}
		
		}
        \caption{Evolutionary parameters for GP}
	\end{table}



		
			
			
           

		

\section{E. Descriptions of Benchmark Datasets}

Each benchmark suite contains ground truth expressions, each paired with a set of numerical data points. As defined in Table~\ref{Dataset_description1}, $U(a, b, c)$ denotes $c$ uniformly sampled random points in $[a, b]$, and $E(a, b, c)$ denotes $c$ equally spaced points in the same range. Here, $a$ and $b$ define the input bounds, and $c$ is the total number of samples. We use $75\%$ of these points for training and the remaining $25\%$ for testing.

\begin{table*}[h]
			
		\scriptsize	\centering
			\renewcommand\arraystretch{0.9}
			\resizebox*{0.9\linewidth}{!}{
				\begin{tabular}{c|c|c|c}
					\toprule
					
					\multicolumn{2}{c|}{\textbf{Benchmark}} & \textbf{Expression} & \textbf{Range} \\ \hline
					
					\multirow{12}{*}{Nguyen} & Nguyen-1 & $x_1^3+x_1^2+x_1$ & $U(-1,1,200)$\\ 
					& Nguyen-2 & $x_1^4+x_1^3+x_1^2+x_1$ & $U(-1,1,200)$ \\ 
					& Nguyen-3 & $x_1^5+x_1^4+x_1^3+x_1^2+x_1$ & $U(-1,1,200)$ \\ 
					& Nguyen-4 & $x_1^6+x_1^5+x_1^4+x_1^3+x_1^2+x_1$ & $U(-1,1,200)$\\
					& Nguyen-5 & $\sin \left(x_1^2\right) \cos (x_1)-1$ & $U(-1,1,200)$ \\ 
					& Nguyen-6 & $\sin (x_1)+\sin \left(x_1+x_1^2\right)$ & $U(-1,1,200)$ \\ 
					& Nguyen-7 & $\log (x_1+1)+\log \left(x_1^2+1\right)$ & $U(0,2,200)$ \\ 
					& Nguyen-8 & $\sqrt{x_1}$  & $U(0,4,200)$\\ 
					& Nguyen-9 & $\sin (x_1)+\sin \left(x_2^2\right)$  & $U(0,1,200)$\\ 
					& Nguyen-10 & $2 \sin (x_1) \cos (x_2)$  & $U(0,1,200)$ \\ 
					& Nguyen-11 & $x_1^{x_2}$  & $U(0,1,200)$ \\ 
					& Nguyen-12 & $x_1^4-x_1^3+\frac{1}{2} x_2^2-x_2$  & $U(0,1,200)$ \\ \hline

					\multirow{6}{*}{Jin} & Jin-1 & $2.5 x_1^4-1.3 x_1^3+0.5 x_2^2-1.7 x_2$ & $U(-3,3,100)$ \\ 
					& Jin-2 & $8.0 x_1^2+8.0 x_2^3-15.0$ & $U(-3,3,100)$\\ 
					& Jin-3 & $0.2 x_1^3+0.5 x_2^3-1.2 x_2-0.5 x_1$ & $U(-3,3,100)$\\ 
					& Jin-4 & $1.5 \exp (x_1)+5.0 \cos (x_2)$ & $U(-3,3,100)$\\ 
					& Jin-5 & $6.0 \sin (x_1) \cos (x_2)$ & $U(-3,3,100)$\\ 
					& Jin-6 & $1.35 x_1 \cdot x_2+5.5 \sin ((x_1-1.0)(x_2-1.0))$ & $U(-3,3,100)$\\  \hline

					\multirow{8}{*}{Constant} & Constant-1 & $3.39x_1^3 + 2.12x_1^2 + 1.78x_1
					$ & $U(-1,1,200)$\\ 
					& Constant-2 & $\sin(x_1^2) \cdot \cos(x_1) - 0.75$ & $U(-1,1,200)$\\ 
					& Constant-3 & $\sin(1.5x_1) \cos(0.5x_2)$ & $U(0,1,200)$\\ 
					& Constant-4 & $2.7x_1^{x_2}$ & $U(0,1,200)$ \\ 
					& Constant-5 & $\sqrt{1.23x_1}$ & $U(0,4,200)$ \\ 
					& Constant-6 & $x_1^{0.426}$ & $U(0,4,200)$\\ 
					& Constant-7 & $2\sin(1.3x_1)\cos(x_2)$ & $U(0,1,200)$\\ 
					& Constant-8 & $\log(x_1 + 1.4) + \log(x_1^2 + 1.3)$ & $U(0,2,200)$\\   \hline
					
					\multirow{22}{*}{Livermore} & Livermore-1 & $\frac{1}{3}+x_1+\sin \left(x_1^2\right)$ & $U(-1,1,200)$ \\ 
					& Livermore-2 & $\sin \left(x_1^2\right) \cos (x_1)-2$ & $U(-1,1,200)$ \\ 
					& Livermore-3 & $\sin \left(x_1^3\right) \cos \left(x_1^2\right)-1$ & $U(-1,1,200)$\\ 
					& Livermore-4 & $\log (x_1+1)+\log \left(x_1^2+1\right)+\log (x_1)$ & $U(0,2,200)$\\ 
					& Livermore-5 & $x_1^4-x_1^3+x_1^2-x_2$ & $U(0,1,200)$\\
					& Livermore-6 & $4 x_1^4+3 x_1^3+2 x_1^2+x_1$ & $U(-1,1,200)$\\ 
					& Livermore-7 & $\sinh (x_1)$ & $U(-1,1,200)$\\ 
					& Livermore-8 & $\cosh (x_1)$ & $U(-1,1,200)$\\   
					&  Livermore-9  & $x_1^9+x_1^8+x_1^7+x_1^6+x_1^5+x_1^4+x_1^3+x_1^2+x_1$ & $U(-1,1,200)$\\ 
					& Livermore-10 & $6 \sin (x_1) \cos (x_2)$ & $U(0,1,200)$\\ 
					& Livermore-11 & $\frac{x_1^2 x_1^2}{x_1+x_2}$ & $U(-1,1,500)$\\ 
					& Livermore-12 & $\frac{x_1^5}{y^3}$ & $U(-1,1,500)$\\ 
					 & Livermore-13 & $x_1^{\frac{1}{3}}$ & $U(0,4,200)$\\ 
					& Livermore-14 & $x^3+x^2+x+\sin (x)+\sin \left(x^2\right)$ & $U(-1,1,200)$\\
					& Livermore-15 & $x_1^{\frac{1}{5}}$ & $U(0,4,200)$\\
					& Livermore-16 &  $x^{\frac{2}{5}}$ & $U(0,4,200)$\\
					& Livermore-17 & $4 \sin (x_1) \cos (x_2)$ & $U(0,1,200)$\\ 
					& Livermore-18 & $\sin \left(x_1^2\right) \cos (x_1)-5$ & $U(-1,1,200)$\\ 
					& Livermore-19 & $x_1^5+x_1^4+x_1^2+x_1$  & $U(0,2,200)$\\ 
					& Livermore-20 &  $\exp \left(-x_1^2\right)$ & $U(-1,1,200)$\\ 
					& Livermore-21 & $x_1^8+x_1^7+x_1^6+x_1^5+x_1^4+x_1^3+x_1^2+x_1$ & $U(-1,1,200)$\\ 
					& Livermore-22 & $\exp \left(-0.5 x_1^2\right)$ & $U(-1,1,200)$\\ 
					\hline

				\end{tabular}
			}
            
			\caption{Benchmark symbolic regression problem specifications}
            
			\label{Dataset_description1}
		\end{table*}

\end{document}